\documentclass[conference]{IEEEtran}

\usepackage{makeidx}  
\makeindex
\usepackage{multirow}
\usepackage[english]{babel}
\usepackage{alltt}
\usepackage[utf8]{inputenc}
\usepackage{amsmath}
\usepackage{amssymb}
\usepackage{graphicx} 
\usepackage{txfonts}
\DeclareGraphicsExtensions{.jpg,.png}
\usepackage{color}
\usepackage{wasysym}
\usepackage{url}
\usepackage[ruled,vlined]{algorithm2e}
\usepackage{eurosym}
\usepackage{color}
\usepackage{rotating}
\usepackage[table,xcdraw]{xcolor}
\usepackage{wrapfig}
\usepackage{caption}
\usepackage{lipsum}
\usepackage{multirow}
\usepackage{rotating}

\makeatletter

\newtheorem{definition}{Definition}

\newcommand{\dotminus}{\mathbin{\text{\@dotminus}}}

\newcommand{\@dotminus}{%
  \ooalign{\hidewidth\raise1ex\hbox{.}\hidewidth\cr$\m@th-$\cr}%
}
\makeatother

\begin{document}

\title{MILP for the Multi-objective VM Reassignment Problem}




\author{
    \IEEEauthorblockN{Takfarinas Saber\IEEEauthorrefmark{1}, Anthony Ventresque\IEEEauthorrefmark{1}\IEEEauthorrefmark{2}, 
Joao Marques-Silva\IEEEauthorrefmark{3}, 
James Thorburn\IEEEauthorrefmark{4}, Liam~Murphy\IEEEauthorrefmark{1}
    }
    \IEEEauthorblockA{\IEEEauthorrefmark{1}Lero@UCD, School of Computer Science and Informatics, University College Dublin, Ireland\\ Email: takfarinas.saber@ucdconnect.ie,    \{anthony.ventresque, 
liam.murphy\}@ucd.ie
    }
    \IEEEauthorblockA{\IEEEauthorrefmark{2}IBM Research, Damastown Industrial Estate, Dublin, Ireland
    }
    \IEEEauthorblockA{\IEEEauthorrefmark{3}Instituto Superior Técnico, Lisboa, Portugal, Email: jpms@ist.utl.pt
    }
    \IEEEauthorblockA{\IEEEauthorrefmark{4}IBM Software Group, Toronto, Canada, Email: jthorbur@ca.ibm.com
    }
}

\maketitle



\begin{abstract}
Machine Reassignment is a challenging problem for constraint programming (CP) and mixed integer linear programming (MILP) approaches, especially given the size of data centres. The multi-objective version of the Machine Reassignment Problem is even more challenging and it seems unlikely for CP or MILP to obtain good results in this context. As a result, the first approaches to address this problem have been based on other optimisation methods, including metaheuristics. In this paper we study under which conditions a mixed integer optimisation solver, such as \emph{IBM ILOG CPLEX}, can be used for the Multi-objective Machine Reassignment Problem. We show that it is useful only for small or medium scale data centres and with some relaxations, such as an optimality tolerance gap and a limited number of directions explored in the search space. Building on this study, we also investigate a hybrid approach, feeding a metaheuristic with the results of CPLEX, and we show that the gains are important in terms of quality of the set of Pareto solutions (+126.9\% against the metaheuristic alone and +17.8\% against CPLEX alone) and number of solutions (8.9 times more than CPLEX), while the processing time increases only by 6\% in comparison to CPLEX for execution times larger than 100 seconds. 

\end{abstract}


\section{Introduction}







\subsubsection*{Background and Research Challenge}

Optimisation of data centres, through reassignment of virtual machines (VMs), is largely seen as one of the biggest challenges in data centre management~\cite{beloglazov2012energy}: not only servers are underutilised~\cite{ahmad2015survey} and the potential savings are important, but the problem has a lot of constraints and is difficult to solve~\cite{corradi2014vm}.
Constraint programming (CP) and mixed\footnote{Note that while the constraints can be expressed by binary variables, the objectives (especially the reliability cost, see Section \ref{problemDefinition}) require real variables.} integer programming (MILP) are known to be inefficient for such large scale problems with a limited execution time~\cite{hermenier2011bin,mehta2012comparing}, and usually researchers focus on other optimisation techniques (e.g., local search~\cite{gavranovic2012variable,brandt2012constraint} or greedy algorithms~\cite{gabay:hal-00764957}) or mix CP or MILP with some other optimisation solutions (e.g., local search~\cite{mehta2012comparing}).


In this paper, we address a slightly different and more relevant problem for the industry: \emph{the Multi-objective VM Reassignment Problem}. 
`Optimising a data centre' seems to suggest that there is something like a \emph{best} reassignment of VMs, but in an enterprise there is no best placement: it is all about which objectives are favoured (by whom and when). 
It is not hard to imagine that different capital allocators (CA, the managers of data centres) may have different perspectives on the best way of making the system better. 
For instance some CAs may consider that energy consumption is the most important element, while for others it can be the cost of licensing; or some CAs see the reliability as the key element (for instance if they run critical applications), while other CAs have a strict policy regarding response time and then collocation of VMs.
These preferences, or policies, may evolve or be in competition: for instance if CAs have two policies for their data centre (e.g., ``electricity has to decrease by \textit{x}\%'' and ``management cost has to be limited to \textit{y}\%'') and some reassignment solutions can serve both, which one to favour? 
In this context, CP and MILP do not seem promising approaches, as the search space is large and the constraints hard and complex. As far as we know, the only related work tackles the problem using some other optimisation techniques. One of the challenges here is that execution time is limited: even if the reassignment is done on a monthly or a  quarterly basis, as it often happens, the decision process is complex and CAs cannot wait more than a few hours or days: they verify and modify the solutions to suit their needs before making any decision. 
In this paper, as it is commonly accepted by practitioners and in the literature \cite{mehta2012comparing,googleRoadefChallenge} we use a time limit for the Multi-objective Reassignment Problem.

However, given that CPLEX's solutions are generally of better quality than the ones of other optimisation techniques, and that there are several relaxation mechanisms in CPLEX and the multi-objective problem itself, we study in this paper whether a MILP solver, such as CPLEX, can be used for the Multi-objective Machine Reassignment Problem.

\subsubsection*{Example}

\begin{figure}
\centering
\includegraphics[width=.25\textwidth]{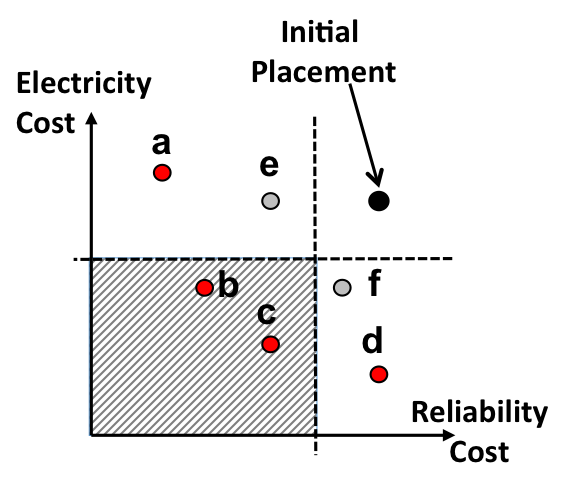}
\caption{\label{motivation} Possible reassignments in 2-dimensions (in red, the good ones) and an organisation's policies (in grey).}
\vspace{-20pt}
\end{figure}


As a motivating example, consider 
Figure \ref{motivation} that shows a reassignment problem with two objectives: the electricity consumption and the reliability (the lower the better for both).
Points a to f are all the possible reassignments, satisfying all the constraints of the data centre.
The good reassignments, i.e., those that are better than any other in a particular combination of objectives, are in red. 
Solution e, for instance, is not a good reassignment as it is worse than b on both objectives - same applies to f wrt. c.
The decision makers may define some policies, i.e., some rules that describe the desired optimisations: here a maximal value for the electricity consumption and the reliability, defining the area of good assignments that the decision makers can select (grey area on Figure~\ref{motivation}). 
There are still two possible reassignments, b and c. The decision makers can then evaluate them locally and make a relative decision among them, such as ``c gives me a better electricity cost but it is not a huge gain, while b's reliability is higher -- so we favour b''. Any technique addressing this problem needs to explore the research space in all directions in order to find varied and good solutions -- which is expensive given the size of the problem.

\subsubsection*{Contributions}

In this paper we make the following contributions: (i) we give a thorough study of how suited is a MILP solver (CPLEX) for the Multi-objective VM Reassignment Problem (the  problem definition is given in Section \ref{problemDefinition}). We show that it is useful only for rather small or medium scale data centres and with some relaxations: a certain tolerance gap and a limited number of directions explored in the search space (Section \ref{CPLEXUsability}); (ii) we propose an algorithm based on the combination of a MILP solver (CPLEX) and a hybrid-metaheuristic to improve the performance of the solver while keeping the execution time acceptable. This is detailed in Section \ref{matheuristic}.

Hybridization is not a novel idea as such (e.g., see \cite{shaw1998using}), however combining a solver (which aim is to find `proper' solutions and not only to relax the problem) and a meta-heuristic to benefit from both worlds in a multi-objective context is novel to our best knowledge. CP and MILP provide better solutions (when they get any, as they are expensive/slow) so our goal here is to help CPLEX as much as we can (i.e., increasing the optimality gap and varying the number of vectors) to get some solutions to feed in the meta-heuristic to cover the search space. 
Note that in \cite{mehta2012comparing} Mehta et al. use CP on a relaxed problem, not the original VM placement one, which leads us to think that CP is inefficient for our problem -- and anyway we also tried ILOG CP Optimizer and noticed extremely poor performances.

\section{Problem Definition}
\label{problemDefinition}

In this section, we give a Multi-objective extension of the VM Reassignment Problem originally proposed by Mehta et al. in \cite{mehta2012comparing}.
We first describe briefly the different elements of data centres, followed by the constraints of the problem, and we finish with the different objectives that we believe are the most relevant (note that this approach is agnostic to objectives and would work with any other linear objective function).

\subsection{Problem Description and Notation}
A data centre is composed of a set $\mathcal{M}$ of physical machines (PMs or servers). Each machine $m_i \in \mathcal{M}$ has a finite amount $Q_{m_i,r}$ of resource $r \in \mathcal{R}$ (e.g., CPU, RAM, storage). Machines $m_i \in \mathcal{M}$ belonging to a same rack are linked with high network connections, and thus considered as being in the same neighbourhood $N(m_i) \in \mathcal{N}$. Resources are of two different types: (i) transient resources ($r\in \mathcal{TR}\subseteq\mathcal{R}$, e.g., RAM or storage) that are consumed at the original host and also at the destination one during a migration process, or (ii) non-transient ($r\in \overline{\mathcal{TR}}$). The data centre is in charge of a set of VMs $v \in \mathcal{V}$ with resource requirements $d_{v, r}$ for every $r \in \mathcal{R}$. $M_0(v)$ and $M(v)$ respectively indicate the initial and the final host of the VM $v$ during the reassignment. VMs composing a same multi-tier application are usually gathered by services $\mathcal{S}=\{s_1, \dots, s_p\}$, with $s_p=\{v_p^1, \dots, v_p^q\}$.

\subsection{Constraints}
We present here the linear constraints of our problem. We follow the problem's linear constraints provided in~\cite{mehta2012comparing}. 

\subsubsection{Reassignment Constraints}
Consider a binary variable $x_{v,m}$ for every VM $v \in \mathcal{V}$ and for each machine $m \in \mathcal{M}$, which is set to 1 if $M(v)=m$ and 0 otherwise. Constraints~(\ref{reassignment}) ensure that every VM is reassigned to one and only one machine: 

\begin{equation}
\forall v \in \mathcal{V},~~~ \sum_{m \in \mathcal{M}} x_{v,m} = 1
\label{reassignment}
\end{equation}

\subsubsection{Capacity Constraints}
There are two ways of computing resource utilisation $U_{m,r}$ of a machine $m \in \mathcal{M}$ and a resource $r \in \mathcal{R}$: (\ref{nonTransient}) for non-transient resources and (\ref{transient}) for transient resources.

\begin{equation}
\forall r \in \overline{\mathcal{TR}},~ \forall m \in \mathcal{M},~~~ U_{m,r} ~~=~~ \sum_{v \in \mathcal{V}} d_{v,r}.x_{v,m}
\label{nonTransient}
\end{equation}

\begin{equation}
\forall r \in \mathcal{TR}, \forall m \in \mathcal{M}, U_{m,r} = \sum_{v \in \mathcal{V} | M_0(v)=m} d_{v,r}.x_{v,m} + \sum_{v \in \mathcal{V} | M_0(v) \neq m} d_{v,r}.x_{v,m}
\label{transient}
\end{equation}

The total resource utilisation of a machine $m \in \mathcal{M}$ should not exceed its capacity $Q_{m,r}$ for every $r \in \mathcal{R}$:

\begin{equation}
\forall r \in \mathcal{R},~ \forall m \in \mathcal{M},~~~ U_{m,r} \le Q_{m,r}
\label{capacity}
\end{equation}

\subsubsection{Conflict Constraints}
VMs which belong to a same service have to be reassigned to different machines (e.g., for replication purposes).

\begin{equation}
\forall s \in \mathcal{S},~ \forall m \in \mathcal{M},~~~ \sum_{v \in s~|~ M(v)=m} x_{v,m} \le 1
\label{conflict}
\end{equation}

\subsubsection{Dependency Constraints}
Services sometimes depend on each other. VMs of these services need to be close to each other in order to achieve high performance (e.g., in the case of multi-tier applications). Let $\mathcal{D}$ be the set of service dependencies such that $\mathcal{D} = \{ (s_i, s_j) | s_i, s_j \in \mathcal{S}$ and $s_i$ depends on $s_j\}$, then:
\begin{equation}
\forall s_i,s_j \in \mathcal{S}, (s_i,s_j) \in \mathcal{D} \implies \forall v_a \in s_i,  \exists v_b \in s_j | N(M(v_a))=N(M(v_b))
\label{depend}
\end{equation}

To give a linear definition of constraints (\ref{depend}), we introduce the binary variables $y_{s,n}$ for every service $s \in \mathcal{S}$ and for each neighbourhood $n \in \mathcal{N}$. Constraints (\ref{dependency1}) and (\ref{dependency2}) ensure that each variable $y_{s,n}$ is set to 1 if at least one VM from the service $s$ is hosted by a machine in the neighbourhood $n \in \mathcal{N}$ and to 0 otherwise:

\begin{equation}
\forall s \in \mathcal{S},~ \forall n \in \mathcal{N}~~~ \sum_{v \in s}\sum_{m \in n}  x_{v,m} \le |\mathcal{N}|.|\mathcal{S}|.y_{s,n}
\label{dependency1}
\end{equation}

\begin{equation}
\forall s \in \mathcal{S},~ \forall n \in \mathcal{N}~~~ \sum_{v \in s}\sum_{m \in n}  x_{v,m} \ge y_{s,n}
\label{dependency2}
\end{equation}

If a service $s_i$ depends on $s_j$, constraints (\ref{dependency3}) guarantee that there is not any VM from $s_i$ reassigned to a machine in a neighbourhood $n \in \mathcal{N}$ without having at least one VM from $s_j$ reassigned to that neighbourhood:
\begin{equation}
\forall (s_i, s_j) \in \mathcal{D},~\forall n \in \mathcal{N} y_{s_i,n} \le y_{s_j,n}
\label{dependency3}
\end{equation}

\subsubsection{Spread Constraints}
Data centres are often decentralised and we assume here that they are composed of different sites $\mathcal{L}$, each machine belonging to a location $l \in \mathcal{L}$. For reliability and security reasons, some services need to be spread over at least \emph{spreadNumber} locations. Let us introduce a binary variable $z_{s,l}$ for every service $s \in \mathcal{S}$ and for each location $l \in \mathcal{L}$. Constraints (\ref{spread1}) and (\ref{spread2}) ensure that $z_{s,l}$ gets the value 1 if the service $s$ has at least one VM running in a machine at the location $l$:

\begin{equation}
\forall s \in \mathcal{S},~ \forall l \in \mathcal{L}~~~ \sum_{v \in s}\sum_{m \in l}  x_{v,m} \le  |\mathcal{N}|.|\mathcal{S}|.z_{s,l}
\label{spread1}
\end{equation}

\begin{equation}
\forall s \in \mathcal{S},~ \forall l \in \mathcal{L}~~~ \sum_{v \in s}\sum_{m \in l}  x_{v,m} \ge z_{s,l}
\label{spread2}
\end{equation}

Constraints (\ref{spread3}) force every service $s$ to run on at least \emph{$spreadNumber_s$} number of locations:
\begin{equation}
\forall s \in \mathcal{S},~~~ \sum_{l \in L} z_{s,l} \ge spreadNumber_s 
\label{spread3}
\end{equation}

\begin{definition}[Machine Reassignment]
\label{def:MachineReassignment}
An assignment $A$ of VMs to machines is a mapping: $A:\mathcal{P} \mapsto \mathcal{M}$, such that $A(v,\mathcal{M})\to m$, which satisfies the constraints (\ref{reassignment}--\ref{conflict} and \ref{dependency1}--\ref{spread3}).\\
A reassignment is a function: $ReA:A\mapsto A$ which returns a new assignment for a given initial assignment of VMs to machines.
\end{definition}

\subsection{Objectives}

We focus here on three objectives: electricity cost, VM migration cost and reliability cost, as they are recognised in the literature~\cite{intelPowerManagment,buyya2009,schroeder2010large} and make sense in practice. The multi-objective variant of the Machine Reassignment (see Definition \ref{def:MachineReassignment}) consists of minimising the cost functions defined by the objectives, the ones we present here or any others that would be relevant.


\subsubsection{Reliability Cost}

For each machine $m \in \mathcal{M}$ and each resource $r \in \mathcal{R}$, we define the safety capacity $SC_{m,r}$ as the amount of resource that it is `safe' to allocate without overloading $m$ -- this is similar to the resource buffer that placement algorithms often assume \cite{xi2015}.
The risk of failure $R_{m,r}$ is then the difference between the actual utilisation of resource $r$ and the safety capacity, and the reliability cost $R_{cost}$ represents the `non reliability' over the full data centre. 


\begin{equation}
\forall m \in \mathcal{M},~\forall r \in \mathcal{R}, ~~~ R_{m,r} \ge U_{m,r} - SC_{m,r}  \ge 0
\label{reliability1}
\end{equation}



\begin{equation}
R_{cost} = \sum_{r \in R} \sum_{m \in M} R_{m,r}
\label{Rcost}
\end{equation}

\subsubsection{Electricity Cost}

For each machine $m \in \mathcal{M}$, we introduce a binary variable $o_m$ to be set by constraints (\ref{machineOn1}) to 1 if the machine $m$ is switched on (i.e., hosts at least one VM) and 0 otherwise.

\begin{equation}
\forall m \in \mathcal{M},~~~  o_m \le \sum_{v \in V} x_{v,m} \le |\mathcal{V}| . o_m 
\label{machineOn1}
\end{equation}


The electricity cost $E_{cost}$ is composed of the electricity consumption of each machine $m \in \mathcal{M}$ multiplied by its price $\gamma_m$. The electricity consumption of a machine $m$ is often modelled as a linear function of its CPU utilisation~\cite{Xu2010moMAP,electricityEstimation2007}, with $\alpha_m$ being its electricity consumption at idle and $\beta_m$ its consumption per unit of CPU usage. 


\begin{equation}
E_{cost} = \sum_{m \in M} \gamma_m \left(\alpha_m . o_m ~~+~~ \beta_m . U_{m, CPU} \right)
\label{Ecost}
\end{equation}

\subsubsection{Migration Cost}
For each VM $v \in \mathcal{VM}$, we introduce a binary variable $mig_v$ to be set by constraints (\ref{mig1}) to 1 if v is reassigned and 0 otherwise.


\begin{equation}
\forall v \in \mathcal{V}, \sum_{m \in M \backslash M_0(v)} x_{v,m} ~=~ mig_v
\label{mig1}
\end{equation}

The migration cost $M_{cost}$ concerns all migrating VMs. For each migrating VM $v \in \mathcal{V}$, the migration cost is the time it takes to: (i) prepare the VM $\mu_1(v)$, (ii) to transfer its image from its initial placement to its new host $m$ $\mu_2(v, M_0(v), m)$ and (iii) to deploy it in the new host $\mu_2(v)$:

\begin{equation}
M_{cost} = \sum_{v \in V} \left(  \left[\mu_1(v) + \mu_3(v)\right] . mig_v ~~+~~ \sum_{m \in \mathcal{M}} \mu_2(v, M_0(v), m) . x_{v,m} \right)
\label{Mcost}
\end{equation}








\section{Experimental Setup}
\label{ExperimentalSetup}

In the current section we present our evaluation setup, i.e., the description of several instances of the problem (data centres to optimise) inspired from a well-known data set~\cite{googleRoadefChallenge} and the metrics used to judge the proposed systems.
All the algorithms described below have been developed in C++. Experiments are done on a computing cluster with 24 cores 2.0GHz Intel Ivy Bridge CPU and 128GB of RAM.

\subsection{Data set}
Google and the ROADEF society (i.e., the French OR society) released a few years ago a data set, now widely used in the OR community, for the evaluation of VM reassignment solutions \cite{googleRoadefChallenge}. This data set represents various data centres, of different sizes and characteristics (e.g., various number of resources), with a large number of constraints. 
This data set does not provide a multi-objective formulation though and we had to adapt the instances (note that the participants of the challenge optimised only one single weighted sum of the costs proposed -- hence there is no possible comparison between our work and others using ROADEF). 
Our instances aim to model realistic scenarios as we observe them in large companies.
For our evaluation, we take the 14 first instances, leaving out only the largest ones (see Table \ref{fig_instances}). 
Two of the objectives we define are present in ROADEF as cost functions: safety/reliability and migration, and we add electricity. 
We randomly generate electricity consumption constants ($\alpha$, $\beta$) for every machine $m \in \mathcal{M}$ and also the electricity cost $\gamma$ for every location $l \in \mathcal{L}$.
The data set also comes with a time limit representing the maximum time allowed for the resolution of the instance (300 seconds). 
We changed the time limit to: 30s for $a\_1\_1$, 1h for $a\_1\_\{2-5\}$, 2h for $a\_2\_\{1-3\}$, and 10h for the other instances, which is considered realistic by large companies in the context of optimisation performed in a regular basis, e.g., monthly or quarterly. 

\begin{table}[!ht]
\centering
\begin{tabular}{| l | c | c | c | c |}
\hline
Instance & \# Resources & \# Machines & \# Services & \# VMs\\
\hline
a\_1\_1 &2     &4       &79     &100   \\
a\_1\_2 &4     &100     &980    &1,000   \\
a\_1\_3 &3     &100     &216    &1,000\\
a\_1\_4 &3     &50      &142    &1,000   \\
a\_1\_5 &4     &12      &981    &1,000   \\
a\_2\_1 &3     &100     &1,000   &1,000   \\
a\_2\_2 &12    &100     &170    &1,000    \\
a\_2\_3 &12    &100     &129    &1,000 \\
a\_2\_4 &12    &50      &180    &1,000 \\
a\_2\_5 &12    &50      &153    &1,000 \\
b\_1    &12    &100     &2,512   &5,000 \\
b\_2    &12    &100     &2,462   &5,000 \\
b\_3    &6     &100     &15,025  &20,000\\
b\_4    &6     &500     &1,732   &20,000  \\

\hline
\end{tabular}

\caption{Characteristics of the different instances used in our evaluation.}
\label{fig_instances}

\vspace{-25pt}
\end{table}

\subsection{Metrics}
The comparison of algorithms optimising a multi-objective problem is complex as their results can be evaluated from different perspectives, such as: the number of solutions on the Pareto front, the variety of solutions~\cite{zitzler2002quality}, or the spread of these solutions. The comparison is even more complex in our case as the problem is large, the Pareto front is not known in most cases and objectives cannot be considered separately. In this paper, we only consider unary metrics, i.e., returning a single value based on solutions found by every algorithm, making a comparison of several algorithms easier.

\subsubsection{Number of solutions}

\begin{figure}
  \centering
    \includegraphics[width=.3\textwidth]{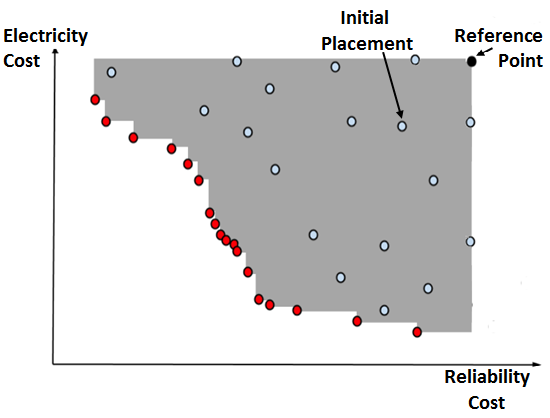}
  \caption{\label{hypervolume}Metrics: number of Pareto solutions (red dots) and hypervolume (grey area).}
\vspace{-25pt}
\end{figure}

We use the number of \emph{non-dominated (efficient) solutions} in our experiments as our first metric. We refer to it as the quantity of found solutions. This metric is highly important for data centre capital allocators as it gives them more choices. It also  provides them with backup solutions if the preferred one appears difficult or technically impossible to implement.

\subsubsection{Hypervolume}

The \emph{hypervolume}~\cite{zitzler1999hypervolume} (a.k.a., $\mathcal{S}$ metric) refers to the quality of a set of solutions. The hypervolume is used~\cite{bringmann2013approximation} in the multi-objective optimisation community for comparing different sets of non-dominated solutions. For every set of solutions, this metric measures the space between the efficient solutions and a reference point far from them. The reference point is defined as a point in the space having the worst objective values and must be identical for all the algorithms, but may be different for every instance. Figure \ref{hypervolume} shows a two-dimensional reassignment, with the non-dominated solutions in red and the other (not interesting) solutions in light blue. The initial placement is one of the solutions, generally not on the Pareto front. The hypervolume is the grey area in this 2D space.

\section{CPLEX for the VM Reassignment Problem}
\label{CPLEXUsability}
The goal of this section is to study both performance and scalability of a MILP solver (i.e., CPLEX) on the Multi-objective VM Reassignment Problem. First, we explore how CPLEX performs on the (Mono-objective) VM Reassignment Problem, i.e., the original problem from ROADEF. We show how difficult the problem is for CPLEX, even in this simpler version. Then we explore the performance of CPLEX for the Multi-objective VM Reassignment and show that it can be tackled under some restrictions, such as, a~limited number of directions (\emph{vectors}) explored in the search space and an optimality tolerance gap.

\subsection{CPLEX for the Mono-objective VM Reassignment Problem}
\label{CPLEXVsMonoVMRP}

Portal et al.~\cite{portal2012simulated} show that the VM Reassignment Problem is too difficult for a MILP solver like CPLEX when applied on the ROADEF instances. CPLEX could only solve 3 instances out of the 14 within the time limit (300s) fixed during the challenge. However CPLEX allows defining an \emph{optimality gap tolerance}, a trigger that stops CPLEX when the current feasible solution falls within a certain percent of CPLEX best estimation of the lower bound (a value smaller or equal to the actual optimal value). For instance, a 5\% gap means that any solution that is 5\% away from the estimated lower bound is accepted and stops CPLEX. In addition, we have here a larger time budget to solve the instances.



\begin{table*}[!ht]
\centering

\scalebox{1}{

\begin{tabular}{l|c|c|c|c|c|c|c|c|c||c|}
\cline{2-11}
\textbf{}                                  & \textbf{50\%} & \textbf{20\%} & \textbf{10\%} & \textbf{5\%} & \textbf{1\%} & \textbf{0.5\%} & \textbf{0.1\%} & \textbf{0.05\%} & \textbf{0.01\%} & \textbf{Limit (s)} \\ \hline
\multicolumn{1}{|l|}{\textbf{a\_1\_1}}     & 0.08 &0.08 & 0.08 & 0.08 & 0.08 & 0.12 & 0.14 & 0.23 & 0.25                      & 30              \\ \hline
\multicolumn{1}{|l|}{\textbf{a\_1\_2}}     & 186              & 183              & 185              & 185             & 1,490            & --            &  --               &  --               & --                &  3,600            \\ \hline
\multicolumn{1}{|l|}{\textbf{a\_1\_3}}     & 27               & 27               & 27               & 37              & 625             & 1,691            &    --             &    --             &    --             &  3,600            \\ \hline
\multicolumn{1}{|l|}{\textbf{a\_1\_4}}     & 50 & 50 & 51 & 51 & 98 & 1,682          & --                &  --               &  --               & 3,600            \\ \hline
\multicolumn{1}{|l|}{\textbf{a\_1\_5}}     & 9                & 9                & 9                & 9               & 9               & 9               & 10              & 19              & 26              &  3,600            \\ \hline
\multicolumn{1}{|l|}{\textbf{a\_2\_1}}     & 54               & 55               & 159              & 3,670            &    --             &        --         &    --             &    --             &    --             &  7,200            \\ \hline
\multicolumn{1}{|l|}{\textbf{a\_2\_2}}     & 2,511             & 2,580             & 2,580             & 2,736            &     --            & --                & --                &     --            & --                &  7,200            \\ \hline
\multicolumn{1}{|l|}{\textbf{a\_2\_3}}     & 71               & 71               & 71               & 71              & 5,816            &  --               &      --           &  --               & --                &  7,200            \\ \hline
\multicolumn{1}{|l|}{\textbf{a\_2\_4}}     & 20,445            & 20,502            & 20,655            &    --             &    --             &    --             &    --             &    --             &    --             & 36,000           \\ \hline
\multicolumn{1}{|l|}{\textbf{a\_2\_5}}     & 21,877            & 22,492            & 22,513            &    --             &        --         &    --             &    --             &    --             &    --             &  36,000           \\ \hline
\multicolumn{1}{|l|}{\textbf{b\_1}}        & 3,482             & 6,913             & 6,913             & 7,094            &     --            & --                &     --            &     --            &     --                      & 36,000           \\ \hline
\multicolumn{1}{|c|}{\textbf{b\_\{2,3,4\}}} &  --                &  --                &     --             &    --             &    --             &    --             &        --         &    --             &        --         &                  36,000           \\ \hline
\end{tabular}

}
\caption{Execution time (s) of CPLEX for the resolution of the identity vector (all objectives have weights equal 1) depending on the optimality tolerance gap. The symbol `--' means that no solution was found in the time limit.}
\label{execTimeCPLEXDifferentGap}
\vspace{-25pt}
\end{table*}

Table \ref{execTimeCPLEXDifferentGap} shows the execution time (in seconds) of CPLEX for the resolution of one single vector (the identity vector, i.e., with weights equal 1 for the three objectives). As a reminder, we also add the time limit (last column) that we set for each instance.
First, we notice that CPLEX only solves instances $a\_1\_1$ and $a\_1\_5$ in the time limit (note that 0.01\% is the default tolerance gap for CPLEX). This supports the general claim that a MILP solver is inefficient for our problem, even in this simple case with only one vector.
We then notice that CPLEX gets a solution with a gap of 0.5\% for all small instances, 10\% for all medium instances (1\% or 5\% for some), and solved only one of the hard instances ($b\_1$, with a 5\% gap), even with a 50\% gap. We also observe that often CPLEX finds a first good solution (e.g., $a\_1\_2$, $a\_1\_3$ and $a\_1\_4$ have a solution for 5\% quickly, as evidenced by the same time for 50\%, 20\%, 10\% and 5\%) but it is then difficult for CPLEX to improve it.
As a conclusion, CPLEX does not seem able to scale to large instances but with a proper gap CPLEX finds good solutions. 





\subsection{CPLEX for the Multi-objective VM Reassignment Problem}
\label{CPLEXVsMultiVMRP}

Once we know that CPLEX finds it difficult to solve one weight vector in the search space, we would like to explore what needs to be relaxed in order to help CPLEX optimise more vectors and hence address a proper multi-objective problem.
In the current section we look at three elements: (i) given that optimising one weight vector is already difficult to CPLEX and the more vectors we optimise the more we explore the search space, what is the most reasonable number of weight vectors to be optimised? (ii) getting a tiny optimality gap increases exponentially the execution time, therefore what is the best value for this parameter? (iii) what is the best way to use CPLEX? Would it be better to collect all intermediate (feasible) solutions found by CPLEX instead of only keeping the best ones for each weight vector?

\begin{figure}[!h]
\centering
\begin{tabular}{c c}

    \includegraphics[width=0.18\textwidth]{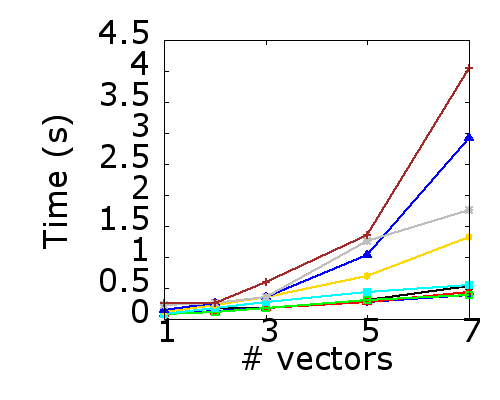} &
    \includegraphics[width=0.18\textwidth]{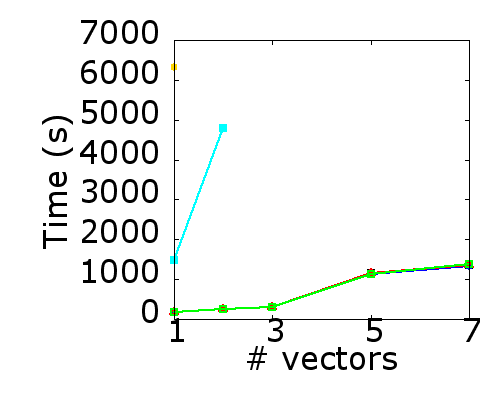} \\
    a\_1\_1 &
    a\_1\_2
    
    \\
    \includegraphics[width=0.18\textwidth]{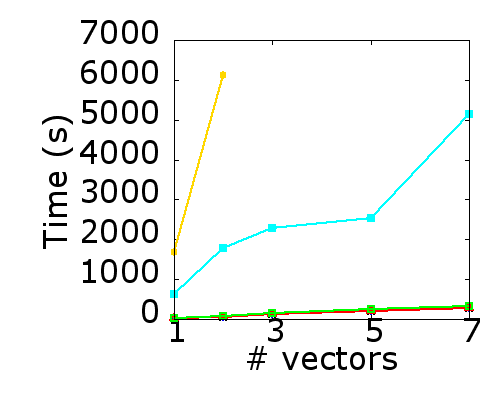} &
    \includegraphics[width=0.18\textwidth]{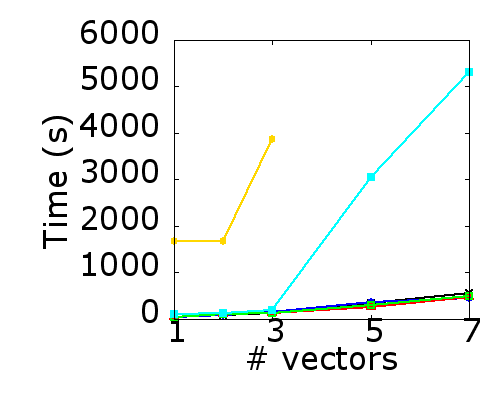} \\
    a\_1\_3 &
    a\_1\_4
    
    \\
    \includegraphics[width=0.18\textwidth]{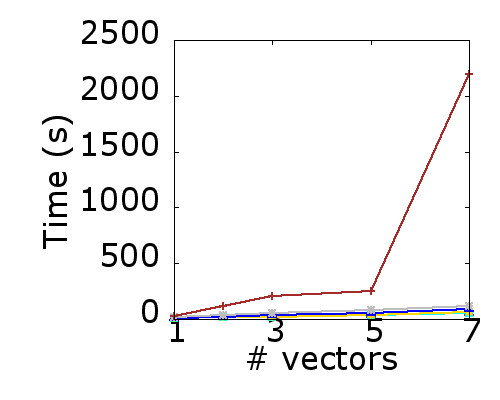} &
    \includegraphics[width=0.18\textwidth]{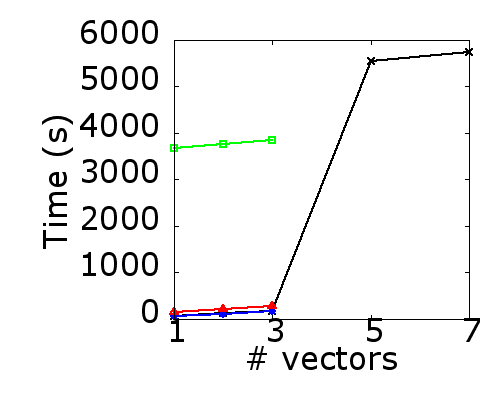} \\
    a\_1\_5 &
    a\_2\_1
    \\

    \includegraphics[width=0.18\textwidth]{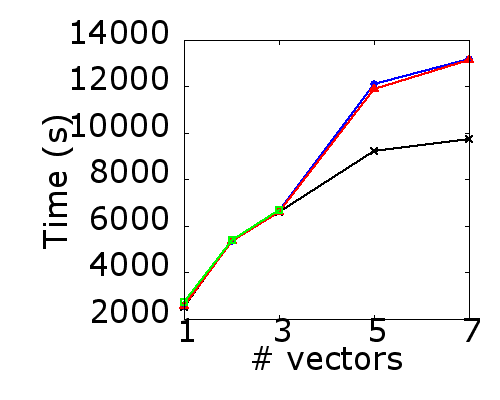} &
    \includegraphics[width=0.18\textwidth]{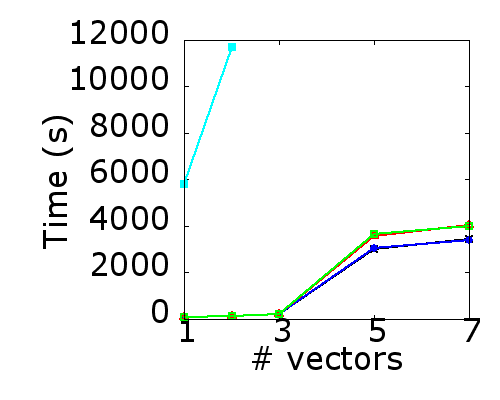} \\
    a\_2\_2 &
    a\_2\_3
    \\
    \includegraphics[width=0.18\textwidth]{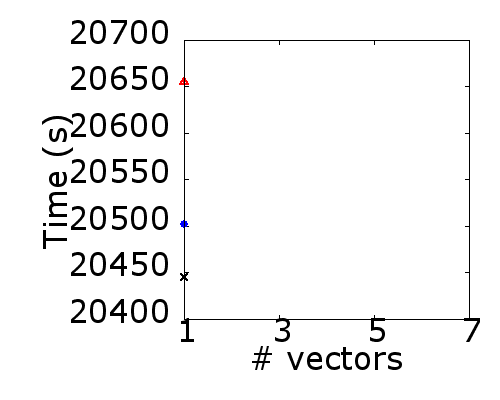} &
    \includegraphics[width=0.18\textwidth]{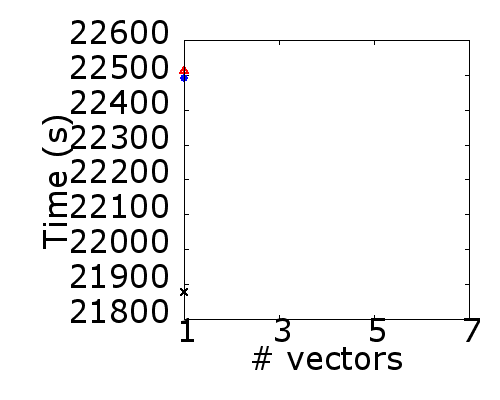} \\
    a\_2\_4&
    a\_2\_5
    \\
    \includegraphics[width=0.18\textwidth]{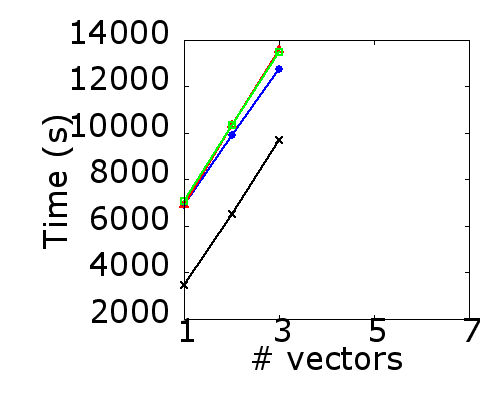} &  
    \includegraphics[width=0.1\textwidth]{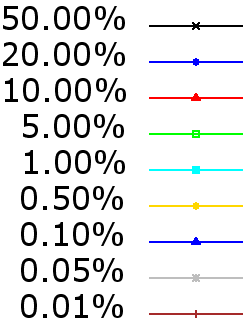}\\
    b\_1 & 
    Optimality Gap
    \\
\end{tabular}
\caption{Execution time (s) of CPLEX on ROADEF instances, with maximally spread objective weight vectors and using different optimality gaps.}
\label{fig:CPLEXMultiIterationsExecutionTime}
\vspace{-25pt}
\end{figure}

    
    

    
    

Figures \ref{fig:CPLEXMultiIterationsExecutionTime} show the execution time CPLEX needs to reach different optimality gaps given several maximally spread weight vectors. These vectors are built on the assumption that to explore a maximum of the space the solver needs to target widely spread directions. In our 3-dimensional space (3 objectives), we successively use these vectors: (1,1,1), (0.6, 0.3, 0.1), (0.3, 0.1, 0.6), (0.1, 0.6, 0.3), (0.45, 0.45, 0.1), (0.45, 0.1, 0.45) and (0.1, 0.45, 0.45). 
We notice that running different vectors increases the final execution time -- as it can be expected. It also confirms what we have seen earlier in Table \ref{execTimeCPLEXDifferentGap} that for large optimality gaps, there is no large difference in terms of execution time (CPLEX quickly finds good solutions), however the more we decrease the gap, the more important is the increase in execution time. We also see that due to the time budget limitation, we cannot run all the possible optimisations with the different vectors for some optimality gaps (e.g., for $a\_1\_2$ we could only use 2 vectors out of the 7 possible ones with an optimality gap of 1\%). Therefore, a decision has to be made on which values should be set for both variables: the optimality gap and the number of vectors. 

According to Figures \ref{fig:CPLEXMultiIterationsExecutionTime}, two patterns emerge: (i) on small instances: asking CPLEX for an optimality gap smaller than 5\% increases its execution time drastically, and (ii) on medium instances: this gap drops to 5 -- 10\%. Therefore, to keep our optimisation in a reasonable execution time, the larger/more complex is the instance the larger the optimality gap we consider. 
Unlike what we might think, CPLEX execution time is very heterogeneous and varies a lot from a vector to another (execution time curves are not linear). Thus, knowing CPLEX execution time on the first vector does not give any indication/prediction on the execution time for other vectors. Because of the lack of knowledge of the execution time, we have to restrict the number of vectors as much as possible. We are even more constrained regarding some instances such as $a\_2\_4$ and $a\_2\_5$ where we could only run CPLEX on one vector.










\begin{figure}[!ht]
\centering
\begin{tabular}{c c c}
& Best Solutions & All Feasible Solutions \\
\begin{turn}{90}~~~~~ a\_1\_1 \end{turn}&
\includegraphics[width=0.18\textwidth]{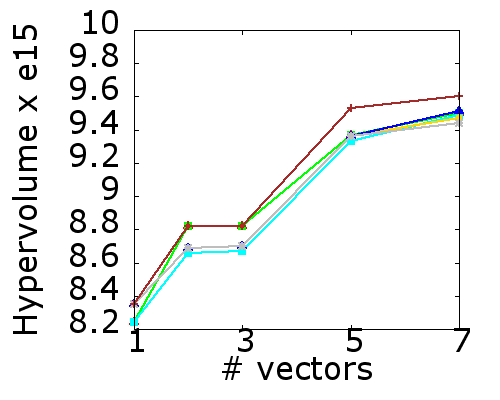}&
\includegraphics[width=0.18\textwidth]{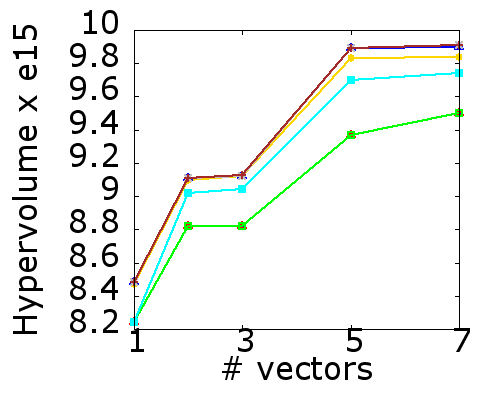}\\
\begin{turn}{90}~~~~~ a\_1\_2 \end{turn}&
\includegraphics[width=0.18\textwidth]{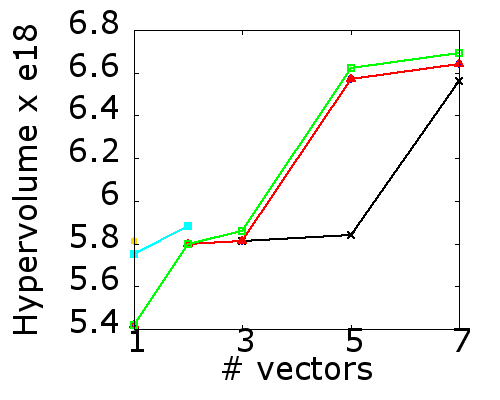}&
\includegraphics[width=0.18\textwidth]{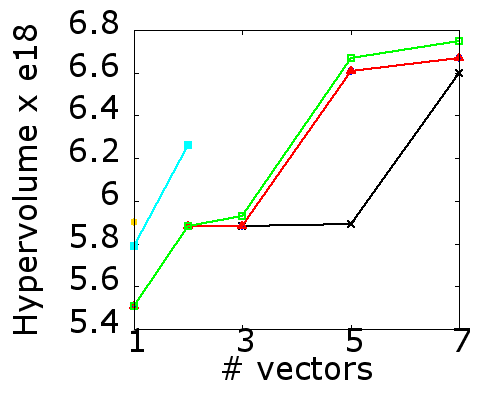}\\
\begin{turn}{90}~~~~~ a\_1\_3 \end{turn}&
\includegraphics[width=0.18\textwidth]{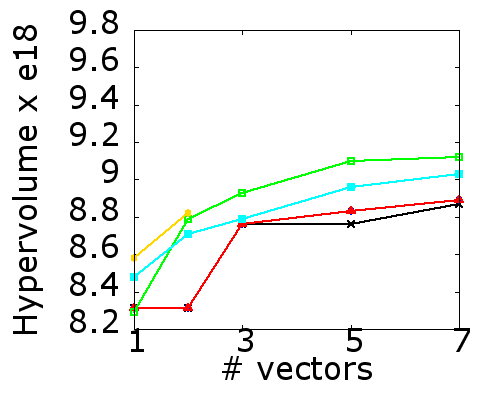}&
\includegraphics[width=0.18\textwidth]{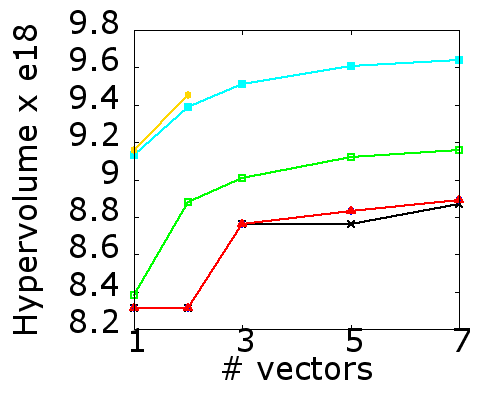}\\
\multicolumn{3}{r}{ \includegraphics[width=0.47\textwidth, height=0.3cm]{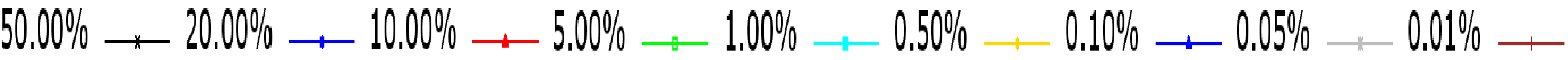}} 
\end{tabular}
\caption{Hypervolume obtained on instances $a\_1\_1$, $a\_1\_2$ and $a\_1\_3$ using CPLEX for different numbers of weight vectors with different optimality gaps by either only considering the best solution found or by collecting all the feasible solutions during the optimisation.}
\label{fig:CPLEXOptimalVsAllFeasible}
\vspace{-25pt}
\end{figure}

Another element of the resolution process that might be interesting to look at is the intermediate (feasible) solutions generated by CPLEX. The way CPLEX works is iterative: it first finds a feasible solution that is either discarded/improved if it is not optimal or kept if it is optimal. 
In our multi-objective context, those intermediate solutions, while not optimal in a particular combination of objectives (remember that CPLEX solves vectors of weights for the objectives), may sometimes carry some interesting reassignments of the VMs, for instance wrt. some single objective -- and hence improve the hypervolume.
Figures \ref{fig:CPLEXOptimalVsAllFeasible} show the hypervolume obtained on the instances $a\_1\_1$, $a\_1\_2$ and $a\_1\_3$ by CPLEX when increasing the number of vectors for different optimality gaps going from 50\% to 0.01\%. The three figures at the left side show the hypervolume obtained when only the best solutions are kept in the solution set while the three figures at the right side give the hypervolume for the same experiments with all solutions in the solution set. 
We notice from the graphs at the left that optimising the objectives using a vector of weights with a small gap does not imply getting a better hypervolume for the multiple objectives (e.g., on the instance $a\_1\_3$, two vectors and a gap of 5\% get a better hypervolume than using a gap of 1\%). This is mainly due to the fact that optimising a compromise of objectives using their linear combination may lead to optimising one objective at the expense of the others. This is also caused by the fact that the objectives are in different units and of different scales (e.g., the electricity cost has a larger scale than the migration cost). 
Collecting all the feasible solutions during the optimisation of every vector may then be a good improvement: see the three graphs at the right of Figure \ref{fig:CPLEXOptimalVsAllFeasible}.
We notice that using a small optimality tolerance gap gives us better results than using a larger one. We also notice that we get an improvement in terms of hypervolume. 
This is an interesting behaviour especially since we did not add any noticeable extra computation time (CPLEX already collects the intermediary solutions, and filtering/removing the dominated solutions requires a negligible execution time). 
In the rest of our evaluations we collect all intermediate (good, i.e., non-dominated) solutions.





\begin{figure}[!ht]
\centering
\begin{tabular}{c c}
    \includegraphics[width=0.18\textwidth]{gnuplotImages2/cplexAlla11.png} &
    \includegraphics[width=0.18\textwidth]{gnuplotImages2/cplexAlla12.png} \\
    a\_1\_1 &
    a\_1\_2\\
    \includegraphics[width=0.18\textwidth]{gnuplotImages2/cplexAlla13.png} &
    \includegraphics[width=0.18\textwidth]{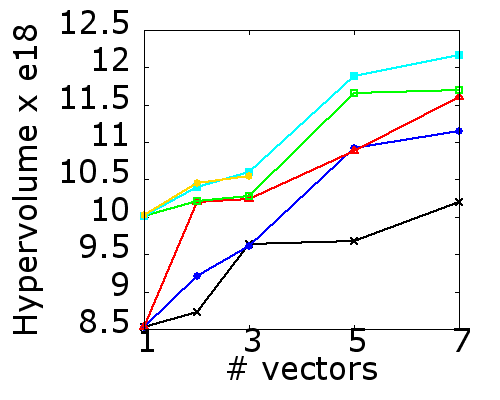} \\
    a\_1\_3 &
    a\_1\_4\\
    \includegraphics[width=0.18\textwidth]{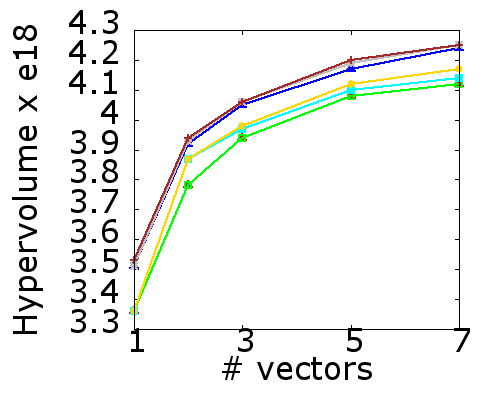} &
    \includegraphics[width=0.18\textwidth]{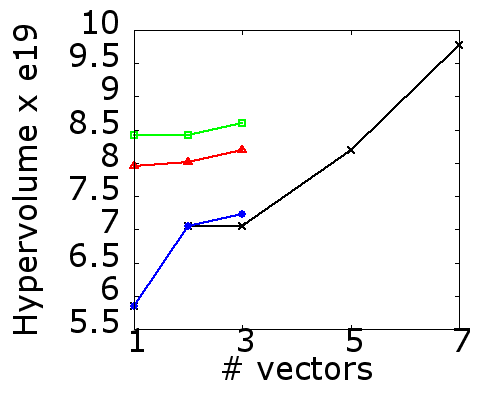} \\
    a\_1\_5 &
    a\_2\_1\\
    \includegraphics[width=0.18\textwidth]{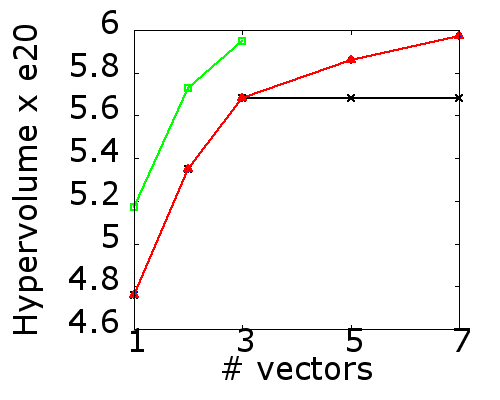} &
    \includegraphics[width=0.18\textwidth]{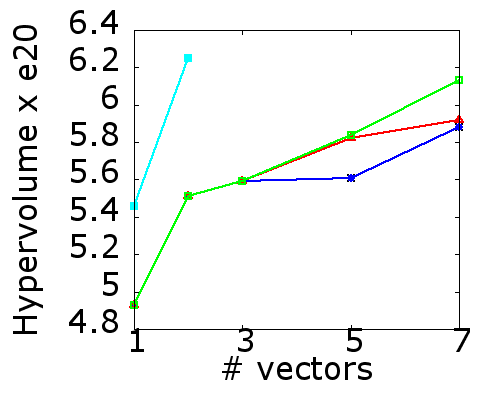} \\
    a\_2\_2 &
    a\_2\_3\\
    \includegraphics[width=0.18\textwidth]{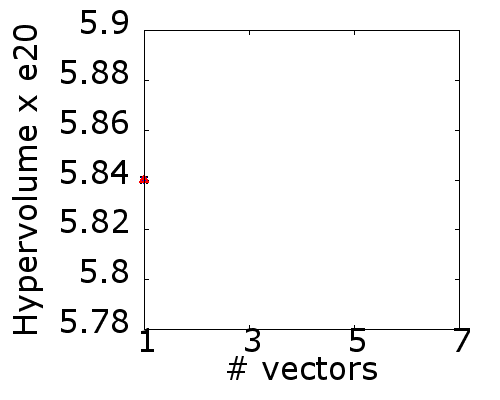} &
    \includegraphics[width=0.18\textwidth]{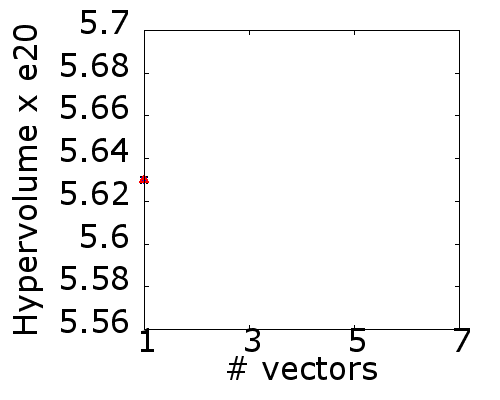} \\
    a\_2\_4&
    a\_2\_5\\
    \includegraphics[width=0.18\textwidth]{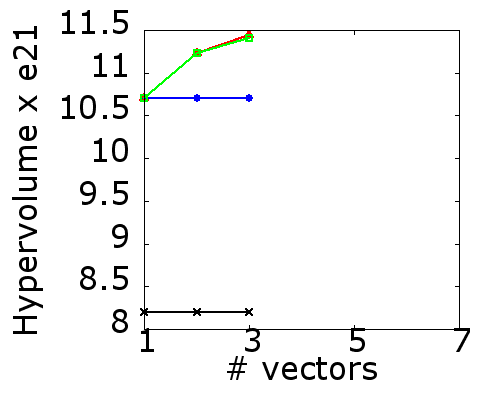} &  
    \includegraphics[width=0.1\textwidth]{gnuplotImages2/legendSquare.png}\\
    b\_1 & 
    Optimality Gap
    \\
\end{tabular}
\caption{Hypervolume obtained with CPLEX (all feasible solutions) on ROADEF instances, using maximally spread weight vectors and different optimality gaps.}
\label{fig:CPLEXMultiIterationsHypervolume}
\vspace{-25pt}
\end{figure}

Figure \ref{fig:CPLEXMultiIterationsHypervolume} shows the hypervolumes obtained with different number of vectors and different tolerance gaps, and this can be read together with Figure \ref{fig:CPLEXMultiIterationsExecutionTime} to figure what composition of number of vectors and tolerance gap gives the best time-hypervolume trade-off.
We notice that globally the hypervolume increases with every new vector. This improvement is more noticeable during the 5 first vectors. It seems to stagnate between the $5^{th}$ and $7^{th}$ (whenever CPLEX reaches them in the time constraint), especially when the optimality gap is tiny. This leads us to think that running several CPLEX optimisations with a large number of vectors (larger than 5) would not be as beneficial as we might think. It would increase the execution time without improving the hypervolume. 
This idea can obviously be withdrawn if the managers of the data centre are ready to spend more time to achieve better hypervolume results.

We see from Figures \ref{fig:CPLEXMultiIterationsExecutionTime} and \ref{fig:CPLEXMultiIterationsHypervolume} that using an optimality gap of 5\% and 3 weight vectors for small instances and an optimality gap of 10\% and 1 weight vector for medium instances allow us to achieve a good hypervolume while keeping the execution time reasonable for all the instances.

\section{Combining a Solver and a Metaheuristic}
\label{matheuristic}

We saw earlier that CPLEX gets good results on small and medium instances, but, it becomes really hard to improve those results without increasing either the number of vectors or the optimality gap, and thus dramatically impacting the execution time. 
In this section, we want to see whether combining a solver with a metaheuristic allows us to get better results in a reasonable time. A comparison of different metaheuristics has already been performed~\cite{saber2014genepi} and a scalable hybrid-metaheuristic (GeNePi) proposed to optimise the Multi-objective VM Reassignment Problem. GeNePi outperforms state-of-the-art algorithms on both quantity and quality of solutions. GeNePi successively applies three metaheuristics: (i) a greedy algorithm (i.e., GRASP~\cite{gabay:hal-00764957}) to quickly find good solutions representing the research space, (ii) a genetic algorithm (i.e., NSGA-II~\cite{deb2002fast}) by combining the previous solutions to get better solutions optimising different objective trade-offs, and (iii) a local search (i.e., PLS~\cite{alsheddy2010guided}) to refine the Pareto front and find more non-dominated solutions.

We take the previous implementation of CPLEX and we give its results to GeNePi, non-dominated solutions found using CPLEX being the initial population. It might happen that CPLEX does not find enough solutions to fill out an entire population (in our case, a population of size 20). In this case the original greedy algorithm in GeNePi is applied to fill this gap and compensate this lack of solutions. In our implementation, GeNePi applies 10 iterations of its second phase (i.e., NSGA-II) in order to evolve the initial population into a fitter one, by getting better and more scattered solutions (spread over the research space). At the end, GeNePi refines the Pareto front by applying a unique iteration of Pareto Local Search (PLS) on the 10 most isolated solutions. Although this step does not bring a large improvement in terms of hypervolume, it is important as it provides decision-makers with more implementation choices. 
Beside these choices, we use the same parameters as in the GeNePi paper~\cite{saber2014genepi}.

\begin{table*}[!ht]
\centering
\begin{tabular}{cc||c|c|c||c|c|c||c|c|c|}
\cline{3-11}
\textbf{}                               & \textbf{}            & \multicolumn{3}{c||}{\textbf{GeNePi}}          & \multicolumn{3}{c||}{\textbf{CPLEX}}           & \multicolumn{3}{c|}{\textbf{CPLEX + GeNePi}}     \\ \hline
\multicolumn{1}{|c|}{\textbf{Instance}} & \textbf{Initial Hyp} & \textbf{Hyp} & \textbf{\#Sol} & \textbf{Time} & \textbf{Hyp} & \textbf{\#Sol} & \textbf{Time} & \textbf{Hyp} & \textbf{\#Sol} & \textbf{Time} \\ \hline
\multicolumn{1}{|c|}{\textbf{a\_1\_1}}  & 6.16E+15             & 9.71E+15     & 225            & 1.58          & 8.82E+15     & 3              & 0.18          & 9.74E+15     & 30             & 0.38          \\ \hline
\multicolumn{1}{|c|}{\textbf{a\_1\_2}}  & 4.75E+18             & 5.69E+18     & 160            & 518           & 5.93E+18     & 8              & 303           & 6.17E+18     & 42             & 319           \\ \hline
\multicolumn{1}{|c|}{\textbf{a\_1\_3}}  & 6.57E+18             & 7.22E+18     & 132            & 441           & 9.01E+18     & 5              & 151           & 9.06E+18     & 21             & 167           \\ \hline
\multicolumn{1}{|c|}{\textbf{a\_1\_4}}  & 8.10E+18             & 9.19E+18     & 136            & 309           & 1.03E+19     & 6              & 138           & 1.09E+19     & 35             & 158           \\ \hline
\multicolumn{1}{|c|}{\textbf{a\_1\_5}}  & 2.42E+18             & 3.15E+18     & 283            & 332           & 3.94E+18     & 4              & 21            & 4.04E+18     & 54             & 34            \\ \hline
\multicolumn{1}{|c|}{\textbf{a\_2\_1}}  & 4.57E+19             & 5.89E+19     & 231            & 284           & 7.95E+19     & 5              & 159           & 8.17E+19     & 39             & 166           \\ \hline
\multicolumn{1}{|c|}{\textbf{a\_2\_2}}  & 3.33E+20             & 4.36E+20     & 197            & 600           & 4.76E+20     & 7              & 2,580          & 5.01E+20     & 42             & 2,694          \\ \hline
\multicolumn{1}{|c|}{\textbf{a\_2\_3}}  & 2.55E+20             & 3.69E+20     & 202            & 695           & 4.93E+20     & 2              & 71            & 5.13E+20     & 32             & 154           \\ \hline
\multicolumn{1}{|c|}{\textbf{a\_2\_4}}  & 3.21E+20             & 6.19E+20     & 253            & 342           & 5.84E+20     & 3              & 20,655         & 6.68E+20     & 43             & 21,173         \\ \hline
\multicolumn{1}{|c|}{\textbf{a\_2\_5}}  & 4.96E+20             & 5.81E+20     & 220            & 347           & 5.63E+20     & 2              & 22,513         & 5.86E+20     & 23             & 22,548         \\ \hline
\multicolumn{1}{|c|}{\textbf{b\_1}}     & 8.20E+21             & 8.74E+21     & 244            & 14,991         & 1.07E+22     & 2              & 8,913          & 1.09E+22     & 29             & 9,452          \\ \hline
\end{tabular}

\caption{Summary of results obtained with GeNePi, CPLEX and CPLEX combined with GeNePi using a gap of 5\% and 3 vectors for small instances, and a gap of 10\% and a unique vector for medium and large instances.}

\label{resultsSummary}
\vspace{-20pt}
\end{table*}

Table \ref{resultsSummary} shows the results obtained on the modified ROADEF instances from $a\_1\_1$ to $b\_1$, in terms of hypervolume, number of non-dominated solutions and execution time. Results are obtained using 10 runs of GeNePi alone (the average is taken), CPLEX alone, and our new matheuristic (i.e., CPLEX combined with GeNePi) while respectively defining the optimality gap  and the number of weight vectors to 5\% and 3 for small instances (i.e., $a\_1\_x$), and 10\% and 1 for medium and large ones (i.e., $a\_2\_x$ and $b\_1$). It also shows the execution time of each algorithm for every instance.
Table \ref{resultsSummary} confirms that GeNePi succeeds in improving the hypervolume and getting a large number of non-dominated solutions while keeping the execution time relatively low. 
We also notice that CPLEX outperforms GeNePi in terms of hypervolume in 8 instances out of 11 with an average improvement of 102\%. However, GeNePi gets on average 63 times more non-dominated solutions (note that the number of solutions while interesting for the decision makers, is not as important as the hypervolume, and anyway too many solutions makes them difficult to explore). We see that CPLEX is slightly better in terms of execution time (given its parallel implementation and the fact that it runs on more powerful memory intensive machines). However we see that CPLEX struggles to scale to large instances (GeNePi gets a better hypervolume on $a\_2\_4$ and $a\_2\_5$ with an execution time of respectively 342s and 347s vs. 21,173s and 22,548s for CPLEX). Compared to CPLEX, we clearly see that adding GeNePi to CPLEX helps to improve the hypervolume (an increase of 17.84\% on average), and also to get more non-dominated solutions (8.9 times more solutions on average), while keeping the execution time low (an average execution time increase of 31.10\%, but of only 6\% for execution times larger than 100s). We also see that CPLEX + GeNePi outperforms the hypervolume obtained by GeNePi alone (with an average increase of 126.96\%) and that unlike CPLEX alone, CPLEX + GeNePi gets better hypervolumes than GeNePi in all instances. CPLEX + GeNePi also gets a fairly reasonable number of non-dominated solutions. However, GeNePi still gets a larger number of solutions (5 times more on average).

To summarise, we can say that CPLEX is good at getting few solutions with a good quality but does not scale well to large instances,  unlike GeNePi and other (hybrid-) metaheuristics.
Also, knowing that CPLEX execution time on one vector is low does not provide much information on the time to run on several other vectors.  
Combining CPLEX with GeNePi seems to be a good solution to improve CPLEX's results in both hypervolume and number of non-dominated solutions with a relatively low increase in the execution time. 
Another advantage of our approach is that it can be adapted to the size of the problem: when the size increases, we have the option to either continue relaxing some of the parameters used for CPLEX (i.e., less vectors or a larger optimality gap) or to replace CPLEX with another `bootstrapping’ technique (e.g., greedy algorithm) to feed in the meta-heuristic. 

\section{Conclusion}
\label{conclusion}

This paper assesses whether a MILP solver, such as CPLEX, can be applied to the Multi-objective VM Reassignment Problem, a large and difficult problem with a lot of complex constraints: given some reassignment objectives (e.g., electricity cost, migration cost, reliability cost), find the best set of reassignment solutions in a limited time -- the limit being quite large (10 hours for large instances).
We show that CPLEX can be used with some relaxations: allowing an optimality tolerance gap (which stops CPLEX when the solutions found are close to the optimal) and limiting the number of directions explored in the search space (giving CPLEX only few vectors of weighted objectives to explore).
We also propose to combine the results of CPLEX to the execution of a metaheuristic (GeNePi) and we compare CPLEX alone, the metaheuristic alone and the combination of both.
We observe that CPLEX is better than the metaheuristic (an improvement of the hypervolume of 102\% in comparison to GeNePi) while the combination CPLEX+GeNePi outperforms both in terms of hypervolume (an average increase of 126.9\% vs. GeNePi and 17.8\% vs. CPLEX), while the execution time remains acceptable (an increase of only 6\% on average in comparison to CPLEX for execution times larger than 100s). 
As future work, we would like to study the usage of algorithms combining metaheuristics and a mathematical resolution for very large instances as a scalable substitution for the MILP solver.

\subsubsection*{{\bf\emph{Acknowledgement}}}
This work was supported, in part, by Science Foundation Ireland grants 10/CE/I1855 and 13/RC/2094, and by Science Foundation Ireland Industry Fellowship grant 13/IF/12789. 

\bibliographystyle{ieeetr}
\bibliography{main}

\end{document}